\documentclass[runningheads]{llncs}

\newif\ifdraft
\drafttrue

\newcommand{\sout}[1]{\ifdraft{\sout{#1}}\else{\vspace{0ex}}\fi}

\usepackage[lined,boxed,linesnumbered]{algorithm2e}

\usepackage{amsmath}
  
\usepackage{cite}
  
\usepackage[english]{babel}
  
\usepackage{multibib}

\usepackage{multicol}  
  
\usepackage{multirow}  
  
\usepackage{pifont}
  
\usepackage{rotating}  
  
\usepackage{soul}  
  
\usepackage[table]{xcolor}

\usepackage[utf8]{inputenc} 
  
\usepackage[normalem]{ulem} 

\usepackage{url} 
  
\newcites{prim}{Primary sources}
\newcites{sec}{References}

\DeclareMathOperator{\aae}{AE}

\DeclareMathOperator{\rae}{RAE}

\newcommand{\killpunct}[1]{} 

\newcolumntype{.}{D{.}{.}{-1}}

\newcolumntype{M}[1]{>{\centering\arraybackslash}m{#1}}

\newcolumntype{N}{@{}m{0pt}@{}}

\newcolumntype{C}[1]{>{\centering\let\newline\\\arraybackslash\hspace{0pt}}m{#1}}

\newcolumntype{Y}{>{\centering\arraybackslash}X}

\begin{document}

\title{LeQua@CLEF2022: Learning to Quantify}

\author{Andrea Esuli, Alejandro Moreo, Fabrizio Sebastiani}
\institute{Istituto di Scienza e Tecnologie dell'Informazione \\
Consiglio Nazionale delle Ricerche \\
56124 Pisa, Italy \\
Email: \emph{firstname.lastname}@isti.cnr.it}


\maketitle

\begin{abstract}
  LeQua 2022 is a new lab for the evaluation of methods for ``learning
  to quantify'' in textual datasets, i.e., for training predictors of
  the relative frequencies of the classes of interest in sets of
  unlabelled textual documents. While these predictions could be
  easily achieved by first classifying all documents via a text
  classifier and then counting the numbers of documents assigned to
  the classes, a growing body of literature has shown this approach to
  be suboptimal, and has proposed better methods. The goal of this lab
  is to provide a setting for the comparative evaluation of methods
  for learning to quantify, both in the binary setting and in the
  single-label multiclass setting. For each such setting we provide
  data either in ready-made vector form or in raw document form.
\end{abstract}


\section{Learning to Quantify}

\noindent In a number of applications involving classification, the
final goal is not determining which class (or classes) individual
unlabelled items belong to, but estimating the \emph{prevalence} (or
``relative frequency'', or ``prior probability'', or ``prior'') of
each class in the unlabelled data. Estimating class prevalence values
for unlabelled data via supervised learning is known as \emph{learning
to quantify} (LQ) (or \emph{quantification}, or \emph{supervised
prevalence estimation}) \cite{Coz:2021xe,Gonzalez:2017it}.

LQ has several applications in fields (such as the social sciences,
political science, market research, epidemiology, and ecological
modelling) which are inherently interested in characterising
\emph{aggregations} of individuals, rather than the individuals
themselves; disciplines like the ones above are usually \emph{not}
interested in finding the needle in the haystack, but in
characterising the haystack. For instance, in most applications of
tweet sentiment classification we are not concerned with estimating
the true class (e.g., \textsf{Positive}, or \textsf{Negative}, or
\textsf{Neutral}) of individual tweets. Rather, we are concerned with
estimating the relative frequency of these classes in the set of
unlabelled tweets under study; or, put in another way, we are
interested in estimating as accurately as possible the true
distribution of tweets across the classes.
  
It is by now well known that performing quantification by classifying
each unlabelled instance and then counting the instances that have
been attributed to the class (the ``classify and count'' method)
usually leads to suboptimal quantification accuracy; this may be seen
as a direct consequence of ``Vapnik's principle'' \cite{Vapnik98},
which states

\begin{quote} If you possess a restricted amount of information for
  solving some problem, try to solve the problem directly and never
  solve a more general problem as an intermediate step. It is possible
  that the available information is sufficient for a direct solution
  but is insufficient for solving a more general intermediate problem.
\end{quote}

\noindent In our case, the problem to be solved directly is
quantification, while the more general intermediate problem is
classification.

Another reason why ``classify and count'' is suboptimal is that many
application scenarios suffer from \emph{distribution shift}, the
phenomenon according to which the distribution across the classes in
the sample (i.e., set) $\sigma$ of \emph{unlabelled} documents may
substantially differ from the distribution across the classes in the
labelled \emph{training} set $L$; distribution shift is one example of
\emph{dataset shift} \cite{Moreno-Torres:2012ay,Quinonero09}, the
phenomenon according to which the joint distributions
$p_{L}(\mathbf{x},y)$ and $p_{\sigma}(\mathbf{x},y)$ differ.  The
presence of distribution shift means that the well-known IID
assumption, on which most learning algorithms for training classifiers
hinge, does not hold. In turn, this means that ``classify and count''
will perform suboptimally on sets of unlabelled items that exhibit
distribution shift with respect to the training set, and that the
higher the amount of shift, the worse we can expect ``classify and
count'' to perform.

As a result of the suboptimality of the ``classify and count'' method,
LQ has slowly evolved as a task in its own right, different (in goals,
methods, techniques, and evaluation measures) from classification.
The research community has investigated methods to correct the biased
prevalence estimates of general-purpose classifiers,
supervised learning methods specially tailored to quantification,
evaluation measures for quantification,
and protocols for carrying out this evaluation.  Specific applications
of LQ have also been investigated, such as sentiment quantification,
quantification in networked environments,
or quantification for data streams.
For the near future it is easy to foresee that the interest in LQ will
increase, due (a) to the increased awareness that ``classify and
count'' is a suboptimal solution when it comes to prevalence
estimation, and (b) to the fact that, with larger and larger
quantities of data becoming available and requiring interpretation, in
more and more scenarios we will only be able to afford to analyse
these data at the aggregate level rather than individually.


\section{The rationale for LeQua 2022}

\noindent The LeQua 2022 lab (\url{https://lequa2022.github.io/}) at
CLEF 2022 has a ``shared task'' format; it is a new lab, in two
important senses:
\begin{itemize}

\item No labs on LQ have been organized before at CLEF conferences.

\item Even outside the CLEF conference series, quantification has
  surfaced only episodically in previous shared tasks. The first such
  shared task was SemEval 2016 Task 4 ``Sentiment Analysis in
  Twitter''~\cite{Nakov:2016ty}, which comprised a \emph{binary
  quantification} subtask and an \emph{ordinal quantification} subtask
  (these two subtasks were offered again in the 2017
  edition). Quantification also featured in the Dialogue Breakdown
  Detection Challenge \cite{Higashinaka:2017cj}, in the Dialogue
  Quality subtasks of the NTCIR-14 Short Text Conversation task
  \cite{Zeng:2019ye}, and in the NTCIR-15 Dialogue Evaluation task
  \cite{Zeng:2020jf}.
  However, quantification was never the real focus of these tasks. For
  instance, the real focus of the tasks described
  in~\cite{Nakov:2016ty} was sentiment analysis on Twitter data, to
  the point that almost all participants in the quantification
  subtasks used the trivial ``classify and count'' method, and
  focused, instead of optimising the quantification component, on
  optimising the sentiment analysis component, or on picking the
  best-performing learner for training the classifiers used by
  ``classify and count''. Similar considerations hold for the tasks
  discussed in \cite{Higashinaka:2017cj,Zeng:2019ye,Zeng:2020jf}.

\end{itemize}
\noindent This is the first time that a shared task whose explicit
focus is quantification is organized. A lab on this topic was thus
sorely needed, because the topic has great applicative potential, and
because a lot of research on this topic has been carried out without
the benefit of the systematic experimental comparisons that only
shared tasks allow.

We expect the quantification community to benefit significantly from
this lab. One of the reasons is that this community is spread across
different fields, as also witnessed by the fact that work on LQ has
been published in a scattered way across different areas, e.g.,
information retrieval
\cite{DaSanMartino:2016jk,Esuli:2018rm,Levin:2017dq}, data mining
\cite{Forman:2008kx,Esuli:2015gh}, machine learning
\cite{Alaiz-Rodriguez:2011fk,Plessis:2017sp}, statistics
\cite{King:2008fk}, or in the areas to which these techniques get
applied \cite{Card:2018pb,Gao:2016uq,Hopkins:2010fk}.  In their own
papers, authors often use as baselines only the algorithms from their
own fields; we thus expect this lab to pull together different
sub-communities, and to generate cross-fertilisation among them.

While quantification is a general-purpose machine learning / data
mining task that can be applied to any type of data, in this lab we
focus on its application to data consisting of textual documents.


\section{Structure of the Lab}


\subsection{Tasks}

\noindent
%
%
Two tasks (T1 and T2) are offered within LeQua 2022, each admitting
two subtasks (A and B).
  
In Task T1 (the \emph{vector task}) participant teams are provided
with vectorial representations of the (training / development / test)
documents. This task has been offered so as to appeal to those
participants who are not into \emph{text} learning, since participants
in this task do not need to deal with text preprocessing
issues. Additionally, this task allows the participants to concentrate
on optimising their quantification methods, rather than spending time
on optimising the process for producing vectorial representations of
the documents.
  
In Task T2 (the \emph{raw documents task}), participant teams are
provided with the raw (training / development / test) documents.  This
task has been offered so as to appeal to those participants who want
to deploy end-to-end systems, or to those who want to also optimise
the process for producing vectorial representations of the documents
(possibly tailored to the quantification task).
  
The two subtasks of both tasks are the \emph{binary quantification
subtask} (T1A and T2A) and the \emph{single-label multiclass
quantification subtask} (T1B and T2B); in both subtasks each document
belongs only to one class $y\in \mathcal{Y}=\{y_{1}, ..., y_{n}\}$,
with $n=2$ in T1A and T2A and $n>2$ in T1B and T2B.
  
\emph{For each subtask in \{T1A,T1B,T2A,T2B\}, participant teams are
not supposed to use (training / development / test) documents other
than those provided for that subtask.} In particular, participants are
not supposed to use any document from either T2A or T2B in order to
solve either T1A or T1B.


\subsection{Evaluation measures and protocols}\label{sec:evaluation}

%
%
In a recent theoretical study on the adequacy of evaluation measures
for the quantification task~\cite{Sebastiani:2020qf}, \emph{absolute
error} (AE) and \emph{relative absolute error} (RAE) have been found
to be the most satisfactory, and are thus the only measures used in
LeQua 2022.  In particular, as a measure we do not use the once widely
used Kullback-Leibler Divergence (KLD), since the same study has found
it to be unsuitable for evaluating quantifiers.\footnote{One reason
why KLD is undesirable is that it penalizes differently
underestimation and overestimation; another is that it is very little
robust to outliers.  See~\cite[\S 4.7 and \S 5.2]{Sebastiani:2020qf}
for a detailed discussion of these and other reasons.}
AE and RAE are defined as
\begin{align}
  \label{eq:ae}
  \aae(p_{\sigma},\hat{p}_{\sigma}) & =\frac{1}{n}\sum_{y\in 
                                      \mathcal{Y}}|\hat{p}_{\sigma}(y)-p_{\sigma}(y)| \\
  \label{eq:rae}
  \rae(p_{\sigma},\hat{p}_{\sigma}) & =\frac{1}{n}\sum_{y\in 
                                      \mathcal{Y}}\displaystyle\frac{|\hat{p}_{\sigma}(y)-p_{\sigma}(y)|}{p_{\sigma}(y)} 
\end{align}
\noindent where $p_{\sigma}$ is the true distribution on sample
$\sigma$, $\hat{p}_{\sigma}$ is the predicted distribution,
$\mathcal{Y}$ is the set of classes of interest, and
$n=|\mathcal{Y}|$.  Note that $\rae$ is undefined when at least one of
the classes $y\in \mathcal{Y}$ is such that its prevalence in the
sample $\sigma$ of unlabelled items is $0$. To solve this problem, in
computing $\rae$ we smooth all $p_{\sigma}(y)$'s and
$\hat{p}_{\sigma}(y)$'s via additive smoothing, i.e., we take
$\underline{p_{\sigma}}(y)=(\epsilon+p_{\sigma}(y))/(\epsilon\cdot n +
\sum_{y\in \mathcal{Y}}p_{\sigma}(y))$,
where $\underline{p_{\sigma}}(y)$ denotes the smoothed version of
$p_{\sigma}(y)$ and the denominator is just a normalising factor (same
for the $\hat{\underline{p_{\sigma}}}(y)$'s); following
\cite{Forman:2008kx}, we use the quantity $\epsilon=1/(2|\sigma|)$ as
the smoothing factor. We then use the smoothed versions of
$p_{\sigma}(y)$ and $\hat{p}_{\sigma}(y)$ in place of their original
non-smoothed versions of Equation~\ref{eq:rae}; as a result, $\rae$ is
now always defined.

As the official measure according to which systems are ranked, we use
RAE; we also compute AE results, but we do not use them for ranking
the systems.

%
%
As the protocol for generating the test samples we adopt the so-called
\textit{artificial prevalence protocol} (APP), which is by now a
standard protocol for the evaluation of quantifiers.  In the APP we
take the test set $U$ of unlabelled items, and extract from it a
number of subsets (the \emph{test samples}), each characterised by a
predetermined vector $(p_{\sigma}(y_{1}), ..., p_{\sigma}(y_{n}))$ of
prevalence values: for extracting a test sample $\sigma$, we generate
a vector of prevalence values, and randomly select documents from $U$
accordingly.\footnote{Everything we say here on how we generate the
test samples also applies to how we generate the development samples.}
%
%
%
%

The goal of the APP is to generate samples characterised by widely
different vectors of prevalence values; this is meant to test the
robustness of a \emph{quantifier} (i.e., of an estimator of class
prevalence values) in confronting class prevalence values possibly
different (or very different) from the ones of the training set.
For doing this we draw the vectors of class prevalence values
uniformly at random from the set of all legitimate such vectors, i.e.,
from the \emph{unit $(n-1)$-simplex} of all vectors
$(p_{\sigma}(y_{1}), ..., p_{\sigma}(y_{n}))$ such that
$p_{\sigma}(y_{i})\in [0,1]$ for all $y_{i}\in\mathcal{Y}$ and
$\sum_{y_{i}\in\mathcal{Y}} p_{\sigma}(y_{i}) = 1$.
For this we use the Kraemer algorithm \cite{smith2004sampling}, whose
goal is that of sampling in such a way that all legitimate class
distributions are picked with equal probability.\footnote{Other
seemingly correct methods, such as drawing $n$ random values uniformly
at random from the interval [0,1] and then normalizing them so that
they sum up to 1, tends to produce a set of samples that is biased
towards the centre of the unit $(n-1)$-simplex, for reasons discussed
in \cite{smith2004sampling}.} For each vector thus picked we randomly
generate a test sample.
We use this method for both the binary case and the multiclass case.

The official score obtained by a given quantifier is the average value 
across all test samples
of RAE, which we use as the official evaluation measure; for each
system we also compute and report the value of AE.  We use the
non-parametric Wilcoxon signed-rank test on related paired samples in
order to assess the statistical significance of the differences in
performance between pairs of methods.


\subsection{Data}

\noindent
The data we use are Amazon product reviews from a large crawl of such
reviews.
From the result of this crawl we remove (a) all reviews shorter than
200 characters and (b) all reviews that have not been recognized as
``useful'' by any users; this yields the dataset $\Omega$ that we will
use for our experimentation.
As for the class labels, (i) for the two binary tasks (T1A and T2A) we
use two \emph{sentiment} labels, i.e., \texttt{Positive} (which
encompasses 4-stars and 5-stars reviews) and \texttt{Negative} (which
encompasses 1-star and 2-stars reviews), while for the two multiclass
tasks (T1B and T2B) we use 28 \emph{topic} labels, representing the
merchandise class the product belongs to (e.g., \texttt{Automotive},
\texttt{Baby}, \texttt{Beauty}).\footnote{The set of 28 topic classes
is flat, i.e., there is no hierarchy defined upon it.}
      
We use the same data (training / development / test sets) for the
binary vector task (T1A) and for the binary raw document task (T2A);
i.e., the former are the vectorized versions of the latter. Same for
T1B and T2B.

The $L_{\mathrm{B}}$ (binary) training set and the $L_{\mathrm{M}}$
(multiclass) training set consist of 5,000 documents and 20,000
documents, respectively, sampled from the
dataset $\Omega$ via \emph{stratified sampling} so as to have
``natural'' prevalence values for all the class labels. (When doing
stratified sampling for the binary ``sentiment-based'' task, we ignore
the topic dimension; and when doing stratified sampling for the
multiclass ``topic-based'' task, we ignore the sentiment dimension).

The development (validation) sets $D_{\mathrm{B}}$ (binary) and
$D_{\mathrm{M}}$ (multiclass) consist of 1,000 development samples of
250 documents each ($D_{\mathrm{B}}$) and 1,000 development samples of
1,000 documents each ($D_{\mathrm{M}}$) generated from
$\Omega\setminus L_{\mathrm{B}}$ and $\Omega\setminus L_{\mathrm{M}}$
via the Kraemer algorithm.

The test sets $U_{\mathrm{B}}$ and $U_{\mathrm{M}}$ consist of 5,000
test samples of 250 documents each ($U_{\mathrm{B}}$) and 5,000 test
samples of 1,000 documents each ($U_{\mathrm{M}}$), generated from
$\Omega\setminus (L_{\mathrm{B}}\cup D_{\mathrm{B}})$ and
$\Omega\setminus (L_{\mathrm{M}}\cup D_{\mathrm{M}})$ via the Kraemer
algorithm.
%
%
A submission for a given subtask will consist of prevalence
estimations for the relevant classes (topic or sentiment) for each
sample in the test set of that subtask.



\subsection{Baselines}

\noindent We have recently developed (and made publicly available)
QuaPy, an open-source, Python-based framework that implements several
learning methods, evaluation measures, parameter optimisation
routines, and evaluation protocols, for LQ
\cite{Moreo:2021bs}.\footnote{\url{https://github.com/HLT-ISTI/QuaPy}}
Among other things, QuaPy contains implementations of the baseline
methods and evaluation measures officially adopted in LeQua
2022.\footnote{Check the branch
\url{https://github.com/HLT-ISTI/QuaPy/tree/lequa2022}}

Participant teams have been informed of the existence of QuaPy, so
that they could use the resources contained in it; the goal was to
guarantee a high average performance level of the participant teams,
since everybody (a) had access to implementations of advanced
quantification methods and (b) was able to test them according to the
same evaluation standards as employed in LeQua 2022.


\section{The LeQua session at CLEF 2022}

\noindent The LeQua 2022 session at the CLEF 2022 conference will host
(a) one invited talk by a prominent scientist, (b) a detailed
presentation by the organisers, overviewing the lab and the results of
the participants, (c) oral presentations by selected participants, and
(d) poster presentations by other participants.

Depending on how successful LeQua 2022 is, we plan to propose a LeQua
edition for CLEF 2023; in that lab we would like to include a
cross-lingual task.


\section*{Acknowledgments}

\noindent This work has been supported by the \textsf{SoBigdata++}
project, funded by the European Commission (Grant 871042) under the
H2020 Programme INFRAIA-2019-1, and by the \textsf{AI4Media} project,
funded by the European Commission (Grant 951911) under the H2020
Programme ICT-48-2020. The authors' opinions do not necessarily
reflect those of the European Commission. We thank Alberto Barron
Cede\~no, Juan José del Coz, Preslav Nakov, and Paolo Rosso, for
advice on how to best set up this lab.



\end{document}